\documentclass{article}
\usepackage{spconf,amsmath,graphicx}

\title{Linear Spatial Pyramid Matching Using Non-convex and non-negative Sparse Coding for Image Classification}
\twoauthors
{Chengqiang Bao, Liangtian He }
{School of Mathematical Sciences \\
University of Electronic Science\\
 and Technology of China\\
Chengdu, Sichuan, 611731\\
 China\\
}
    {Yilun Wang\sthanks{This work was supported by the Natural Science Foundation of China, Grant
Nos. 11201054, 91330201, by  the National Basic Research Program (973 Program), Grant No. 2015CB856000, and by the Fundamental Research Funds for the Central Universities ZYGX2013Z005.}}
	{School of Mathematical Sciences \\
University of Electronic Science\\
and Technology of China\\
Chengdu, Sichuan, 611731 China\\
Center for Information in BioMedicine\\
University of Electronic Science\\
and Technology of China,
Chengdu 610054, China\\
Center for Applied Mathematics\\
Cornell Unviersity\\
Ithaca, NY, 14850 USA\\
E-mail: yilun.wang@gmail.com
}

{}

\usepackage[color,final]{showkeys}

\begin{document}
%
\maketitle
\begin{abstract}

Recently sparse coding have been highly successful in image classification mainly due to its capability of incorporating the sparsity of image representation. In this paper, we propose an improved sparse coding model based on linear spatial pyramid matching(\emph{SPM}) and Scale Invariant Feature Transform (\emph{SIFT} ) descriptors. The novelty is the simultaneous non-convex and non-negative characters added to the sparse coding model.  Our numerical experiments show that the improved approach using non-convex and non-negative sparse coding is superior than the original \emph{ScSPM}\cite{YK09} on several typical databases.
\end{abstract}
\section{Introduction}
\label{sec:intro}

\begin{figure}[t]

\begin{minipage}[b]{1.0\linewidth}
  \centering
  \centerline{\includegraphics[width=8.5cm]{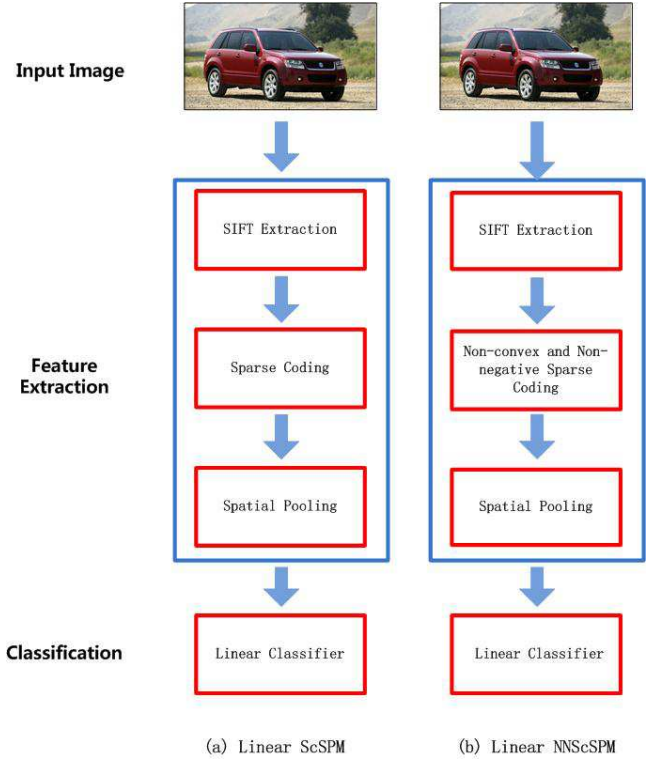}}
\end{minipage}
\caption{Schematic comparison of the original ScSPM with our proposed NNScSPM, which is an improved algorithm based on the former.}
\label{fig:res}
\end{figure}

In recent years, image classification, as an important part of image processing problems, has been a research focus in computer vision. Support Vector Machines (SVMs) using Spatial Pyramid Matching (SPM) have been highly successful in image classification. SPM model, as a feature extraction method, is extended from bag-of-words(BoW) model, which is a document representation method in the field of information retrieval. This model treats documents as some keywords combinations, ignores the syntax and sequence of the texts, finally matches the documents with the frequencies of keywords. Researchers have applied this method onto image processing and turned up  the bag-of-features(BoF), which represents an image as a histogram of its local features.

Unfortunately,  the method is incapable of capturing shapes or locating an object, because of this attribute discarding the spatial order of local features. In order to overcome this issue, spatial pyramid matching (SPM) is proposed, as the most successful extension of the BoF model by incorporating the geometric correspondence search, discriminative codebook learning and the generative part model. The SPM method partitions an image into $2^{l}\times 2^{l}$ segments in different scales $l=0,1,2$, and computes the BoF histogram to form a vector representation of the image. In cases where only the scale $l=0$, SPM reduces to BoF. Due to the utility of the spatial information, SPM method is more efficient than BoF and also has shown very promising performance on many image classification issues.

Figure \ref{fig:res} is a typical flowchart that clearly illustrates the SPM approach. The (a) part is the flow chart of the linear SPM methods.  The flow contains SIFT extraction, sparse coding, spatial pooling and classification. The (b) part is our flowchart that is based on sparse coding model. Firstly, in the descriptor layer, feature points like SIFT descriptors can be extracted from the input image. Then a codebook is applied to quantize each descriptor and obtain the code layer. In the next SPM layer, multiple codes from inside each sub-region are pooled together by averaging and normalizing into a histogram. Finally, the histograms from all sub-regions are concatenated together to generate the final representation of the image for the following classification task.

It has been  empirically found that, traditional SPM has to use classifiers with a particular type of nonlinear Mercer kernels, e.g. the intersection kernel or the Chi-square kernel for achieve good performance. Accordingly, the cost of nonlinear classifiers, bearing $O(n^{3})$ , computational complexity in training and \emph{O(n)} for testing in SVM($n$ is the number of support vectors), are very expensive. So it's impractical for large scale real-world applications.

To settle this issue, Yang et al\cite{YK09} proposed a method called linear spatial pyramid matching using sparse coding(ScSPM), which  achieves state-of-the-art performance on several databases in image categorization experiments. In the following years, some improved methods based on ScSPM are introduced, like \emph{LLC}\cite{WangJ10} and \emph{LR-ScSPM}\cite{Gaos10}, which mainly take into account the features's locality in the feature quantization process based on sparse coding to improve the accuracy of image classification.

In this paper, we  propose an improved image classification algorithm based on ScSPM, which can apply non-convex and non-negative properties admirably to image sparse representation to improve the accuracy of image classification. It is well known that there are two  fundamental steps in image representation. One is coding, where sparse coding is widely used now, and the other is spatial pooling, which can mainly get the feature representation of images. However, traditional sparse coding followed multi-scale max pooling has no constraints on the sign of coding coefficients. In order to avoid the loss of some information loss in the following max pooling, the non-zero components  are conditioned to non-negative in our sparse coding model, as did in \cite{ZhangC11}. Considering that the sparsity of images representations is also essential\cite{YK09,WangJ10,Gaos10},  we employ non-convex algorithm-ISD\cite{Wang09}, in order to  get more sparse representations. In sum, we propose non-convex and non-negative sparse coding and we call the improved ScSPM as  NNScSPM.

The remainder of the paper is organized as follows. Section $2$ introduces the basic idea of of ScSPM and describes our proposed NNScSPM. Section $3$ presents experiment results. Finally, section $4$ concludes our paper.

\section{Linear SPM Using SIFT Non-convex and Non-negative Sparse Codes}
\label{sec:LSPM of use sparse coding}
There are two basic feature extraction methods in image classification: vector quantization and sparse coding. In this section, we introduce these two models firstly, and then we propose our model and algorithm.

\subsection{Vector Quantization (VQ)}
\label{ssec:VQ}

Let $X$ be a set of SIFT appearance descriptors in a $L$ dimensional feature space, i.e. $X=[x_1,x_2,\dots,x_\mathrm{M}]^T\in{R^{M\times{L}}} $.
 The vector quantization(VQ) method applies the K-means algorithm\cite{Has09} to solve the following problem:
\begin{equation}
\label{eq:1}
\quad \min_D  \sum_{m=1}^{M}\min_{l=1\dots p} ||x_m-d_l||^2
\end{equation}
where $D=[d_{1},\dots,d_{p}]^{T}$  are the $p$ cluster centers to be found, called \emph{codebook} or \emph{dictionary}, and $\|\bullet\|$ denotes the $ l_{2}$-norm of vectors. The optimization problem can be rewritten into a matrix factorization problem with cluster membership indicators $A=[\alpha_{1},\dots,\alpha_{M}]^{T}$:
\begin{eqnarray}\label{eq:2}
&min_{A,D}&  \sum_{m=1}^{M} ||x_m-D\alpha_m||^2\\\nonumber
&s.t.& \emph{Card}(\alpha_m)=1,|\alpha_m|=1,\alpha_m\geq0,\forall m
\end{eqnarray}
where $ \emph{Card}(\alpha_m)=1 $ is a cardinality constraint, meaning that only one element of $ \alpha_m $ is nonzero; $\alpha_m\geq0,$  means that all the elements of $\alpha_m $ are nonnegative, and $ |\alpha_m| $ is the $ l_1- $norm of $\alpha_m $, i.e., the summation of the absolute value of each element in $\alpha_m $. After the optimization problem (\ref{eq:2}) is solved, the index of the only nonzero element in $\alpha_m $ indicates which cluster the vector $ x_{m} $ belongs to.

The objective of Eq.(2) is non-convex and the minimization is always alternatively performed  with respect to the labels $ A $ with $D$ fixed, and with respect to $D$ with the labels fixed\cite{Has09}. This phase is called training. After that, the coding phase will be going. The learned $D$ will be applied or tested on a new set of $X$ and in such cases, the problem (\ref{eq:2}) will be solved with respect to $A$ only.

\subsection{Sparse Coding}
\label{ssec:SC}
Because of restrictive the constraint $ \emph{Card}(\alpha_m)=1 $ , which makes the vector $ x_{m} $ responding to only a element of the codebook, the reconstruction of $X$ may be too coarse. To overcome this issue, Kai Yu et al.\cite{YK09} relaxed the constraint by instead putting an $ l_{1}- $norm regularization on $ \alpha_{m} $, which enforces  $ \alpha_{m} $ to have a  small number of nonzero elements. Then the $VQ$ formulation is turned into another problem known as \emph{sparse coding} (SC):
\begin{equation}
\label{eq:3}
 \min_{A,D}  \sum_{m=1}^{M} ||x_m-D\alpha_m||^2+\lambda|\alpha_{m}|
\end{equation}
Where $ \lambda $ is a trade-off parameter for balancing  the fidelity term and the sparse regularization term. In order to avoid trivial solutions, a unit $ l_{2}- $norm constraint on $\alpha_{m} $ is typically applied. Normally, the dictionary $D$ is an overcomplete basis set, i.e. $ P\succ L $. Solving (\ref{eq:3}) is similar to that of the problem (\ref{eq:2}) consisting of  a training phase and coding phase. 

\subsection{Our model and algorithm for Sparse Coding}
\label{ssec:NNSC}

\subsubsection{Our Model: Non-convex and Non-negative Sparse Coding}
\label{sssec:model}

Because of less restrictive constraint, SC coding can achieve a much lower reconstruction error than VQ coding. Research in image statistics has clearly disclosed that natural image patches are sparse signals, and the sparse representation can help capture salient properties of images.

The SC model is based on $ l_{1}-$norm, which is a popular sparsity enforcement regularization due to its convexity. However, the non-convex sparse regularization such as the widely used $ l_{p}-$norm ($ 0\leq p <1 $), prefers an even more sparse solution. Although there have existed many works based on non-convex penalization for image processing, to our best knowledge, there have existed only few specific works for image classification. The major difficulties with the existing non-convex algorithms are that the global optimal solution cannot be efficiently computed for now, the behavior of a local solution is also hard to analyze and more seriously the prior structural information of the solution is hard to be incorporated.

SC model in \cite{YK09} followed by max pooling did not use non-negative constraint. It's well known that in order to satisfy the optimization, negative coefficients are usually needed. Because non-zero components typically reflect  remarkable feature information, max pooling will bring the loss in terms of these negative components, and moreover reduce the accurate of image classification\cite{ZhangC11}.

Taking those two situation account, this paper proposes a non-convex and non-negative sparse model based on the convex sparse coding and proposes a multistage convex relaxation algorithm for it via our proposed iterative support detection\cite{Wang09}, which has proved to be more sparse than the pure $ l_{1} $ solution in theory in many cases. Our new sparse model is a truncated $\ell_1$ model with nonnegative constraints as  follows:
\begin{eqnarray}\label{eq:4}
&min_{A,D}&  \sum_{m=1}^{M}(||x_m-D\alpha_m||^2+\lambda\sum_{k=1}^{p}w_{m_k}|\alpha_{m_k}|)\\\nonumber
&s.t.&  \alpha_m\geq0,\forall m
\end{eqnarray}
where $ w_{m_k} $ are the $0-1$ weights, i.e. the components of $\alpha$ corresponding to $0$ weights are removed out of the $\ell_1$ penalty, and will not be penalized \cite{Wang09}. 

While ones  commonly consider the plain $\ell_1$ model: $ \min_{\alpha} |x-D\alpha||^2+\lambda\sum_{k=1}^{p}|\alpha_{m_k}|$, i.e.,  all the weights $w_{m_k}$ are the same and equal to $1$,  in practice, we might be able to obtain more prior information beyond the sparsity assumption. For example, we might know the locations of certain coefficients $\alpha_{m_k}$ of large nonzero absolute values. In such cases, we should not penalize these coefficients in the $l_1$ norm regularization and remove them out of the $\ell_1$ norm (corresponding $w$ weights being $0$) and use a truncated $\ell_1$ norm \cite{Wang09}.

\subsubsection{Our Algorithm: ISD-YAll1 algorithm}\label{sssec:al}

 The difficulty is that this kind of partial support information is not available beforehand in practice. Correspondingly, in this paper, we propose to take advantage of the idea of the Iterative Support Detection (ISD, for short) in \cite{Wang09} to extract the reliable information about the true solution and set up $0-1$ weights correspondingly. The procedure is an alternating optimization, which repeatedly take the following two steps:

Step 1: Optimize $\alpha$ with $ w_{m_k} $ fixed (initially  $\vec{1}$): this is a convex problem in $\alpha$.

Step 2: Determine the value of $w_{m_k} $ according to the current $\alpha$. The value of weights will be used in Step 1 of the next iteration.

 For Step 1, the truncated nonnative $ \ell_{1}-$norm optimization problems can be efficiently solved via  the Yall1 algorithms \cite{YangJ09}. For Step 2, the locations of  large nonzeros are estimated from the solution of the last (truncated) $\ell_1$-norm optimization problem via the support detection procedure. 
 Our algorithm is named as ISD-Yall1, and described as follows:\\
\\
\begin{tabular}{l}
\hline
\textbf{ Algorithm } The Proposed  ISD-Yall1 Algorithm   \\
\hline
Given X extracted an image and the codebook $D$.\\
1.Set the iteration $itr\leftarrow 0$ and initialize \\
the set of detected entries $I^{(itr)}\leftarrow \emptyset$.\\
2. while the stopping condition is not satisfied,   \\
(a) $w^{itr}$ $\leftarrow$ $(I^{(itr)})^C :=$ $\mathbf{\Omega}$ $ \backslash$ $I^{(itr)}$;\\
(b) $\alpha^{(itr)}$$\leftarrow$ solve truncated $l_1$ minimization  for $\alpha= \alpha^{(itr)}$;\\ (Step 2: using Yall1 method)\\
(c) $I^{(itr+1)}$ $\leftarrow$ support detection using $\alpha^{(itr)}$ as the reference;\\ (Step 1: using \textit{threshold}-ISD strategy)\\
(d) $itr$$\leftarrow$ $itr+1$.\\
\hline
\end{tabular}\vspace{0.2cm}
 Here $\mathbf{\Omega}$ denotes the  universal set of $(i,j,l), 1\leq i\leq M, j\in \{1,2,\ldots,L\}, l\in \mathcal{I}$. The support detection, i.e. the set of the nonzeros of large magnitudes, are estimated as follows \cite{Wang09}.
\begin{equation}\label{eq:threshold ISD}
I^{(itr+1)} :=\{(i,j,l) : |w_{i,j,l}^{itr}| > \epsilon ^{(itr)}\},
\end{equation} $itr=0,1,2,\ldots$

In pooling phase, because of our non-negative model, the pooling function is defined as  $z_j=$max$\{ \alpha_{1j},\alpha_{2j},\dots,\alpha_{Sj}\} $,  where $z_j$ is the j-th element of $z$, which is used in linear SVM classifier for image classification.  $\alpha_{ij} $is the matrix element at i-th row and j-th column of $A$, and $S$ is the number of local descriptors in the region.

\section{experiment results}
\label{sec:exp}
In this section, we report the comparisons results between our proposed non-convex and non-negative model and ScSPM on two widely used public datasets: 15 scenes\cite{15sc} and UIUC-Sport dataset\cite{ui}. Experiments's parameters setting will be analyzed in this section. Besides our own implementations, we also quote some results directly from the literature, especially those of ScSPM from \cite{YK09} and \cite{Gaos10}. All the experiments were performed under Windows 7 and Matlab (R2013b) running on a desktop with an Inter(R)CPU i5-4590(3.3GHZ) and 8GB of memory.

\subsection{Parameters Setting}
\label{ssec:PS}

Local features descriptor is essential to image representation. In our work, we also adopt the widely used SIFT feature due to its excellent performance in image classification. To fairly compare with others, we use 50,000 SIFT descriptors extracted from random patches to train the codebook which is same as \cite{YK09} in the train phase. In sparse coding phase, the most important two parameters are (i): the sparsity regularization parameter  $\lambda$. The performance is best in ScSPM\cite{YK09} when it is 0.2-0.4. We follow the same setting of the interval (0.2,0.4). (ii): the threshold value, which is used for computing the weights of coefficients. we take the threshold as following:$\epsilon ^{(itr)} \doteq \max(\alpha^{itr})/\beta^{itr+1}$, where  the performance is good when $\beta$ is $1.3-1.5$ empirically. In our experiments, we compare our results to ScSPM and KSPM, which uses spatial-pyramid histograms and Chi-square kernels\cite{YK09}. NScSPM is the non-negative sparse model which only use the non-negative constraint, not use the non-convex.

\subsection{15 Scene Data Set}
\label{ssec:15data}

Scene $15$ contains $15$ categories and $4485$ images in all, with $200$ to $400$ images per category. The image content is diverse, containing not only indoor scene, such as bedroom, kitchen, but also outdoor scene, such as buildings and country \textit{etc}. To compare with others¡¯ work, we randomly select 100 images per class as training data and use the rest as test data. The detailed comparison results are shown in the Table 1.

\begin{table}[!h]
\caption{Performance Comparison on 15 Scene Dataset}
\centering
\begin{tabular}{c|c}
\hline
\hline
Method & Average Classification Rate($\%$) \\
\hline
KSPM & 76.73$\pm$0.65\\
ScSPM\cite{YK09} &80.28$\pm$0.93\\
NScSPM & 81.30$\pm$0.53\\
NNScSPM & 81.92$\pm$0.42\\
\hline
\end{tabular}

\end{table}

\subsection{UIUC-Sport Data Set}
\label{ssec:uiuc}
UIUC-Sport contains 8 categories and 1792 images in all, and the image number ranges from 137 to 250. These 8 categories are badminton, bocce, croquet, polo, rock climbing,rowing, sailing and snow boarding. We also randomly select 70 images from each class as training data and use the rest as test data. We list our results in Table 2.

\begin{table}[!h]
\caption{Results on UIUC-Sport Dataset}
\centering
\begin{tabular}{c|c}
\hline
\hline
Method & Average Classification Rate($\%$) \\
\hline
ScSPM &82.85$\pm$0.62\\
NScSPM & 83.53$\pm$0.72\\
NNScSPM & 84.13$\pm$0.37\\
\hline
\end{tabular}

\end{table}

 From the results, we have observed two points that: (i) non-convex and non-negative properties can play an important role on image classification indeed and (ii) our proposed NNScSPM is superior than KSPM and ScSPM on 15 scene dataset and UIUC-Sport dataset.

\section{concludes}
\label{sec:con}

We propose a non-convex and non-negative sparse coding model for image classification in this paper, which is efficiently solved by proposed ISD-Yall1 algorithm. The non-convex property reflects the sparsity of the image and the non-negative property can avoid the loss in max spatial pooling. Our numerical experiments effectively demonstrates its better performance.


%



\begin{thebibliography}{1}

\bibitem{YK09} Yang J, Yu K, Gong Y, et al. Linear spatial pyramid matching using sparse coding for image classification[C]//Computer Vision and Pattern Recognition, 2009. CVPR 2009. IEEE Conference on. IEEE, 2009: 1794-1801.
\bibitem{WangJ10}J. Wang, J. Yang, K. Yu, F. Lv, T. Huang, and Y. Gong. Locality-constrained linear coding for image classification. In CVPR, 2010.
\bibitem{ZhangC11}Chunjie Zhang, ling Liu, Qi Tian, Changsheng Xu, Hanqing Lu, and Songde Ma, "Image classification by non-negative sparse coding, low-rank and sparse decomposition, " in Computer Vision and Pattern Recognition, 2011, pp. 1673-1680.
\bibitem{He14}He L, Wang Y. Iterative Support Detection Based Split Bregman Method for Wavelet Frame Based Image Inpainting.  IEEE Transactions on Image Processing, vol.23, no.12, pp.5470,5485, Dec. 2014
\bibitem{Wang09}Wang Y, Yin W. Sparse signal reconstruction via iterative support detection[J]. SIAM Journal on Imaging Sciences, 2010, 3(3): 462-491.
\bibitem{Gaos10} Gao S, Tsang I W, Chia L T, et al. Local features are not lonely¨CLaplacian sparse coding for image classification[C]//Computer Vision and Pattern Recognition (CVPR), 2010 IEEE Conference on. IEEE, 2010: 3555-3561.
\bibitem{Lee06}Lee H, Battle A, Raina R, et al. Efficient sparse coding algorithms[C]//Advances in neural information processing systems. 2006: 801-808.
\bibitem{XShen12}Shen X, Wu Y. A unified approach to salient object detection via low rank matrix recovery[C]//Computer Vision and Pattern Recognition (CVPR), 2012 IEEE Conference on. IEEE, 2012: 853-860.
\bibitem{ZJ13}Ji Z, Theiler J, Chartrand R, et al. SIFT-based Sparse Coding for Large-scale Visual Recognition[J]. SPIE Defense Security Sens, 2013.
\bibitem{YangJ09}Yang J, Zhang Y. Alternating direction algorithms for $l_1$-problems in compressive sensing, ArXiv e-prints, 2009[J].
\bibitem{Has09}Hastie T, Tibshirani R, Friedman J, et al. The elements of statistical learning[M]. New York: Springer, 2009.
\bibitem{YB10}Boureau Y L, Bach F, LeCun Y, et al. Learning mid-level features for recognition[C]//Computer Vision and Pattern Recognition (CVPR), 2010 IEEE Conference on. IEEE, 2010: 2559-2566.
\bibitem{Bach14}Mairal J, Bach F, Ponce J. Sparse Modeling for Image and Vision Processing[J]. arXiv preprint arXiv:1411.3230, 2014.
\bibitem{LowSift04}Lowe D G. Distinctive image features from scale-invariant keypoints[J]. International journal of computer vision, 2004, 60(2): 91-110.
\bibitem{bagCG04}Csurka G, Dance C, Fan L, et al. Visual categorization with bags of keypoints[C]//Workshop on statistical learning in computer vision, ECCV. 2004, 1(1-22): 1-2.
\bibitem{SJbag03}Sivic J, Zisserman A. Video Google: A text retrieval approach to object matching in videos[C]//Computer Vision, 2003. Proceedings. Ninth IEEE International Conference on. IEEE, 2003: 1470-1477.
\bibitem{YangJ10}Yang J, Yu K, Huang T. Supervised translation-invariant sparse coding[C]//Computer Vision and Pattern Recognition (CVPR), 2010 IEEE Conference on. IEEE, 2010: 3517-3524.
\bibitem{GK05}Grauman K, Darrell T. The pyramid match kernel: Discriminative classification with sets of image features[C]//Computer Vision, 2005. ICCV 2005. Tenth IEEE International Conference on. IEEE, 2005, 2: 1458-1465.
\bibitem{15sc}Lazebnik S, Schmid C, Ponce J. Beyond bags of features: Spatial pyramid matching for recognizing natural scene categories[C]//Computer Vision and Pattern Recognition, 2006 IEEE Computer Society Conference on. IEEE, 2006, 2: 2169-2178.
\bibitem{ui}Li L J, Fei-Fei L. What, where and who? classifying events by scene and object recognition[C]//Computer Vision, 2007. ICCV 2007. IEEE 11th International Conference on. IEEE, 2007: 1-8.

\end{thebibliography}

\end{document}